\documentclass[lettersize,journal]{IEEEtran}
\usepackage{amsmath,amsfonts}
\usepackage{algorithmic}
\usepackage{algorithm}
\usepackage{array}
\usepackage[caption=false,font=normalsize,labelfont=sf,textfont=sf]{subfig}
\usepackage{textcomp}
\usepackage{stfloats}
\usepackage{url}
\usepackage{verbatim}
\usepackage{graphicx}
\usepackage{cite}
\usepackage{hyperref}
\usepackage{multirow}
\usepackage{booktabs}

\begin{document}

\title{Predicting Price Movements in High-Frequency Financial Data with Spiking Neural Networks}


\author{Brian Ezinwoke$^1$, Oliver Rhodes$^1$
\thanks{$^1$Department of Computer Science, University of Manchester, UK}
}

\vspace{-1cm}
\maketitle

\vspace{-1cm}
\begin{abstract}
Modern high-frequency trading (HFT) environments are characterized by sudden price spikes that present both risk and opportunity, but conventional financial models often fail to capture the required fine temporal structure. Spiking Neural Networks (SNNs) offer a biologically inspired framework well-suited to these challenges due to their natural ability to process discrete events and preserve millisecond-scale timing. This work investigates the application of SNNs to high-frequency price-spike forecasting, enhancing performance via robust hyperparameter tuning with Bayesian Optimization (BO). This work converts high-frequency stock data into spike trains and evaluates three architectures: an established unsupervised STDP-trained SNN, a novel SNN with explicit inhibitory competition, and a supervised backpropagation network. BO was driven by a novel objective, Penalized Spike Accuracy (PSA), designed to ensure a network's predicted price spike rate aligns with the empirical rate of price events. Simulated trading demonstrated that models optimized with PSA consistently outperformed their Spike Accuracy (SA)-tuned counterparts and baselines. Specifically, the extended SNN model with PSA achieved the highest cumulative return (76.8\%) in simple backtesting, significantly surpassing the supervised alternative (42.5\% return). These results validate the potential of spiking networks, when robustly tuned with task-specific objectives, for effective price spike forecasting in HFT.

\end{abstract}


\section{Introduction}

High Frequency Trading (HFT) involves automated execution of trades on microsecond timescales, relying on tick data and adhering to extreme {latency} constraints \cite{kohda2021hft, Gao2023}. Financial time series data are inherently difficult due to {non-stationarity} (statistical properties change over time) \cite{KimTaeYoon2004Annf}, which makes models vulnerable to {concept drift} \cite{sezer2020financial}. They also exhibit {volatility clustering} and {heavy-tailed distributions}, indicating frequent extreme events and noise \cite{Chakraborti, Gao2023}. 

{Spiking Neural Networks (SNNs)} offer a compelling solution for this domain \cite{Lobo2020, Yamazaki2022}. They process information using discrete binary activations (spikes), naturally preserving the fine temporal dynamics essential for detecting significant movements in financial time series\cite{Gao2023} . This sparse computation yields inherent {low latency} and {energy efficiency} compared to dense Artificial Neural Networks (ANNs) \cite{tavanaei2018deep, Gao2023, davidson2021digital}, especially on neuromorphic hardware. 

However, a significant hurdle in training SNNs is the lack of a proven learning algorithm capable of extracting these temporal properties in real-world applications. The firing of a spiking neuron is a threshold-based, discontinuous event \cite{Eshraghian2023, Roy2019}, which complicates the direct application of gradient-based optimisation used in ANNs \cite{KozdonThesis2018, Eshraghian2023}. To enable gradient descent training in multi-layer SNNs, {surrogate gradient methods} have become crucial \cite{Roy2019}. These methods replace the discontinuous activation function with a smooth, differentiable surrogate during the backward pass \cite{Eshraghian2023, Roy2019}, allowing weight updates via backpropagation-like mechanisms \cite{Eshraghian2023, Roy2019}. While successful for deep SNNs, these methods lack biological plausibility and fast online learning capabilities.

This work focuses on training models using an unsupervised learning rule based on {Spike-Timing-Dependent-Plasticity (STDP)}, a mechanism where synaptic strength ($\Delta w$) is modified by the relative timing ($\Delta t$) of pre- and postsynaptic spikes \cite{Gerstner1996, Hebb1949, Bi1998}. Specifically, a presynaptic spike preceding a postsynaptic spike causes strengthening (LTP), while the reverse causes weakening (LTD) \cite{KozdonThesis2018}. The magnitude and direction of $\Delta w$ are determined by a learning window, $W(\Delta t)$ \cite{KozdonThesis2018}. A common mathematical formulation of the STDP learning window is the exponential window \cite{Gao2023}:
\begin{equation}
    \Delta w = \eta \cdot  W(\Delta t) =
\begin{cases}
A_{+} \exp(-\Delta t / \tau_{+}) & \text{if } \Delta t > 0 \\
-A_{-} \exp(\Delta t / \tau_{-}) & \text{if } \Delta t < 0 \\
0 & \text{otherwise }
\end{cases}
\label{eq:stdp}
\end{equation}
where $A_{+}$ and $A_{-}$ are the maximum learning rates for potentiation and depression, respectively (with $A_{-} \leq A_{+}$); $\tau_{+}$ and $\tau_{-}$ are the corresponding time constants defining the temporal window; and $\eta$ is a learning rate parameter, typically set to $1$ \cite{KozdonThesis2018}.

STDP-trained SNNs are highly sensitive to their many hyperparameters (Eq.~\ref{eq:stdp} and neuronal dynamics). Therefore, a robust hyperparameter search method is necessary. {Bayesian optimisation (BO)} is a probabilistic model-based approach designed to optimise computationally expensive functions, such as model validation performance \cite{frazier2018bayesian}. Unlike grid or random search, BO uses the results of previous evaluations to inform the choice of the next hyperparameters, efficiently finding the optimal configuration with fewer evaluations \cite{bergstra2013hyperparams}. In this paper, BO is used to optimise model and learning algorithm parameters in tandem.

Reid, Hussain, and Tawfik (2014) introduced {Polychronous Spiking Networks (PSNs)}, leveraging their intrinsic temporal capabilities for general financial time series prediction \cite{Reid2013}. Benchmarking the PSN against MLPs, DRPNNs, and a linear model, they demonstrated favourable results, confirming the PSN paradigm's potential in non-stationary environments, though their approach was supervised \cite{Reid2013}.

Most relevant to this work is the study by Gao et al. (2021), which specifically targets high-frequency price-spike prediction using SNNs \cite{Gao2023}. Addressing the research gap focused on supervised direction prediction, and proposed a novel method based on unsupervised STDP. This approach enabled the network to become selective to temporal patterns in intra-day data, demonstrating SNN potential for identifying specific events such as price spikes. A limitation noted was the need for improved computational efficiency for ultra-low latency HFT requirements \cite{Gao2023}.

To address the challenge of hyperparameter optimisation in SNNs, Parsa et al. (2019) proposed and evaluated a {Bayesian-based hyperparameter optimisation framework} for neuromorphic systems \cite{parsa2019bayesian}. They highlighted that conventional methods such as grid search are inefficient in the high-dimensional parameter space of SNNs. Their iterative Bayesian methodology significantly streamlined the search process, finding optimal parameters with substantially fewer evaluations (e.g., 400 vs. 24,000 for input encoding), underscoring BO's utility for complex SNN architectures \cite{parsa2019bayesian}.

This work explores the continuation and extension of these concepts, particularly STDP-based unsupervised training for HFT predictions. Specifically, the work contributes:
\begin{itemize}
\item Development of an extended SNN integrating temporal features and inhibitory synapses for financial time series prediction.
\item Empirical demonstration that unsupervised STDP-trained networks significantly outperform supervised gradient-based methods and baseline strategies in HFT.
\item Establishment of a fully reproducible, end-to-end training framework utilizing Bayesian optimisation to achieve fine-grained control over network performance, through hyperparameter optimisation.
\item Proposal of a new penalised spike accuracy metric as a target for BO, which was able to outperform the spike accuracy metric proposed by Gao et al.
\end{itemize}

\section{Method}

\subsection{Data and Preprocessing}
\label{subsec:data}

High-frequency \$AAPL stock price and volume data with microsecond precision from the 19 trading days of February 2015 was used for training and evaluating the model. This dataset provides sufficient granularity to capture the rapid price movements characteristic of HFT environments while maintaining a manageable size for computational efficiency.

The raw price time series cannot be input directly to the SNN as it must first be converted into spike trains. A series of preprocessing steps are conducted as follows.

\subsubsection{Volume Weighted Average Prices}
\label{subsubsec:vwap}

The volume weighted average price (VWAP) is used across all models presented in this paper. For a time period with $n$ transactions, VWAP is evaluated according to Eqn.~\ref{eq:vwap}:
\begin{equation}
\text{VWAP} = \frac{\sum_{i=1}^{n} P_i \cdot V_i}{\sum_{i=1}^{n} V_i}
\label{eq:vwap}
\end{equation}
where $P_i$ represents the price of the $i$-th transaction, and $V_i$ represents its corresponding volume. A window length of $n=10$ timestamps to average across  was selected during this study, effectively reducing overall data size by 90\%.

\begin{figure}[!tb]
    \centering\includegraphics[width=0.95\columnwidth]{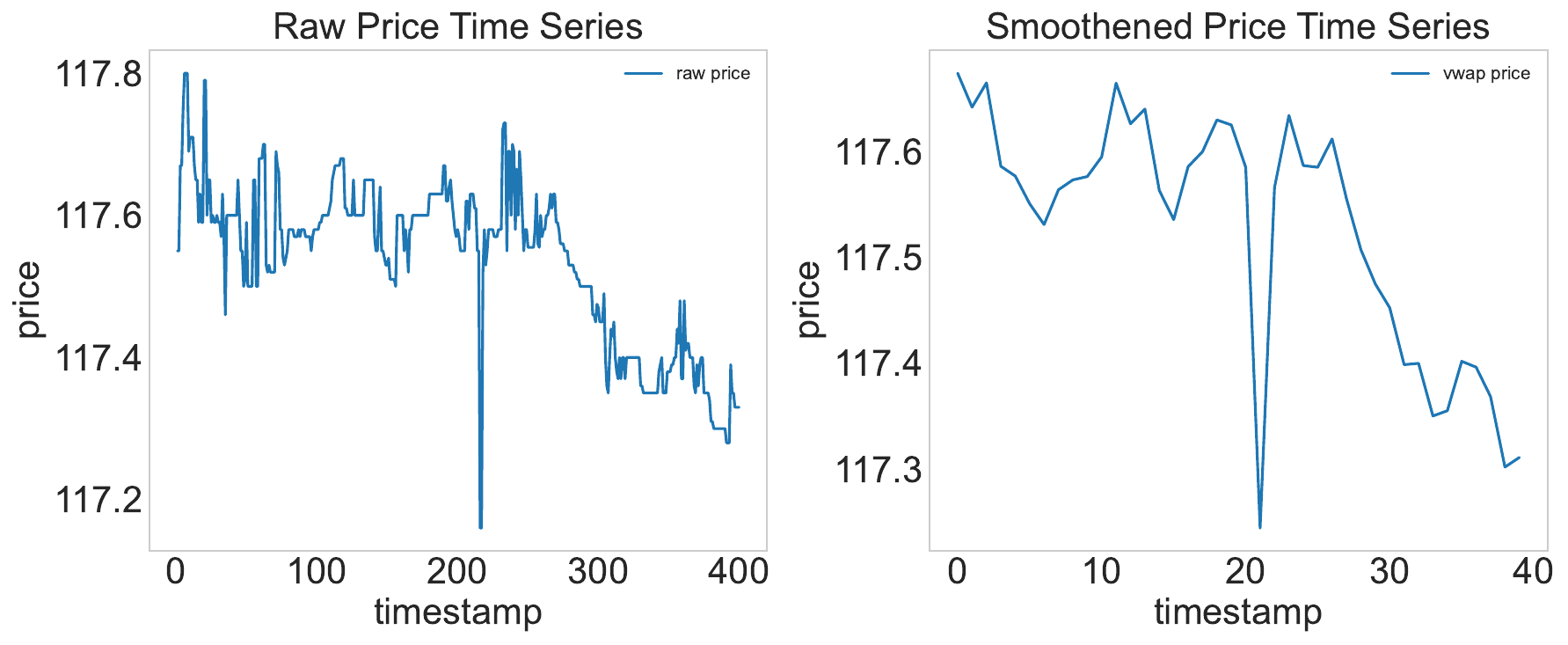} 
    \caption{Comparison of the raw transaction price (left) and the VWAP (right).}
    \label{fig:raw-vwap-comparion}
\end{figure}

The VWAP conversion serves multiple purposes: smooths the data to eliminate zig-zag noise from bid-ask oscillations; reduces the number of data points, improving computational efficiency; and incorporates volume information, which is economically significant as price changes correlate with trading volume \cite{Gao2023}. Figure~\ref{fig:raw-vwap-comparion}, provides a direct comparison of the raw price time series (left) compared to the smoothed VWAP time series (right). The two panels show the price move for the same contract for the same period, yet the VWAP is visibly smoother and the significant price spikes are more obvious. Note that the horizontal-axis of the left panel is 10 times the x-axis of the right panel since a window length 10 timestamps was used to aggregate the raw price into VWAP \cite{Gao2023}.

\subsubsection{Feature Creation}
\label{subsubsec:feature_creation}

\begin{figure}[!htb] 
    \centering
    \includegraphics[width=\columnwidth]{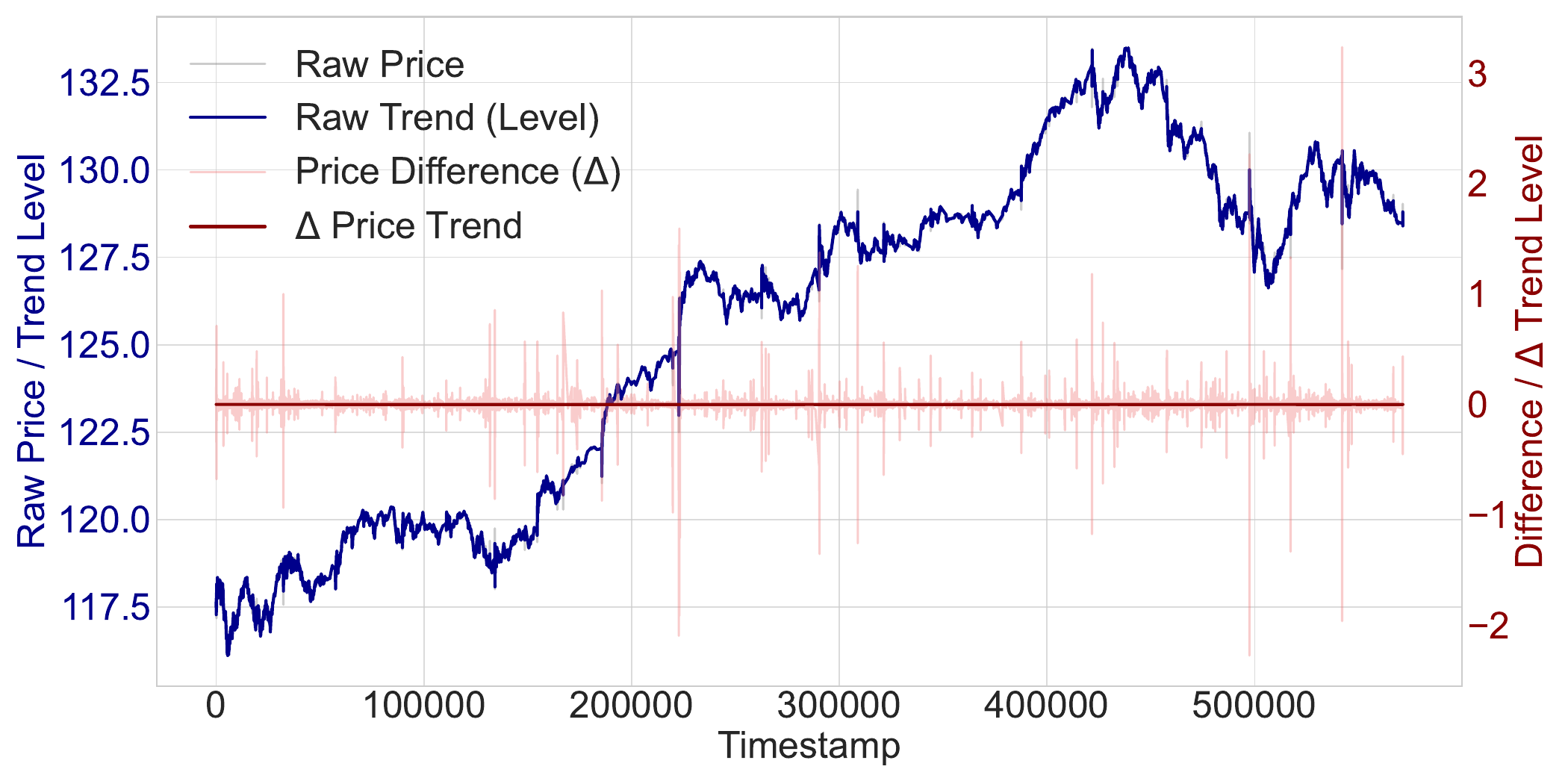}
    \caption{Comparison of trend components in price vs. differenced price series using unobserved component analysis (UCA) \cite{durbin2012timeseries}\cite{seabold2010statsmodels}.}
    \label{fig:trend-comparison}
\end{figure}

The next step is to select the features necessary to train the SNN models; however, feeding the VWAP directly is insufficient for several reasons. First, there exist day-trend components in price time series which can add noise into the SNN and cause disturbance to the detection of price spikes \cite{Gao2023}, as seen in Figure~\ref{fig:trend-comparison}. Second, the effect of the different price magnitudes cannot be ignored. Large differences between the open and close prices would cause the rate of the generated spike trains in the beginning hour of trading to be higher than the rate at the day's end. This bias could lead to a systematic difference in spike frequency for different trading hours, and have a negative influence on the performance of an SNN. 

The proposed solution to these problems is to take the difference of the prices since price spikes are  related only to the price change and not the magnitude \cite{Gao2023}. As seen in Figure~\ref{fig:trend-comparison}, this removes the trend component and prevents systematic time biases from entering the model. Additional features providing more relative information relating to the price change can also be included, given the trend component is removed by differencing or some other method. 

\subsubsection{Normalisation}
\label{subsubsec:normalisation}
Before data encoding into spike trains, features are normalised to the range $[0, 1]$. This range then corresponds to the probability of a spike at a timestamp or the rate parameter for Poisson encoding.

In the unsupervised spike learning method proposed by Gao et~al., data is normalised using a variant of min-max z-score normalisation. A normalised feature $z_i$ is calculated as:
\begin{equation}
    z_i = \max\{ \frac{x_i - \mu_\mathbf{x}}{\sigma_\mathbf{x}} \cdot \sigma_{\text{norm}} + \mu_{\text{norm}}, 0 \}
\end{equation}
where $x_i$ is the $i$-th value in the feature vector $\mathbf{x}$, and $\mu_\mathbf{x}$ and $\sigma_\mathbf{x}$ are the mean and standard deviation of $\mathbf{x}$. The parameters $\mu_{\text{norm}}$ and $\sigma_{\text{norm}}$ are network hyperparameters. This method presents two major challenges.

First, finding optimal $\mu_{\text{norm}}$ and $\sigma_{\text{norm}}$ is computationally expensive, requiring multiple rounds of data encoding and model training. Poor selection of $\mu_{\text{norm}}$ can lead to network over-excitation due to high spiking rates. Likewise, an improper $\sigma_{\text{norm}}$ selection can result in a contracted or overly wide data range, either causing all spike trains to be overly similar or causing similar input values to have drastically different spike counts.

Second, the normalisation is significantly affected by outliers (e.g., abnormally large values). If a large outlier exists, most non-outlier data clusters near the new minimum (zero) after scaling. This distortion adversely impacts the predictive performance of the network by obscuring the true information within the time-series data.

To address these issues, an alternative three-step normalisation approach is adopted:

\begin{enumerate}
  \item \textbf{Robust Scaling \& Clipping:} Apply robust scaling using the interquartile range (IQR) $[0.1, 0.9]$. Values outside this range are clipped to the nearest boundary. This mitigates outlier effects, preventing data clustering near the $[0, 1]$ bounds after subsequent scaling.
 \item \textbf{Channel Separation:} Split the scaled features into positive and negative channels to represent bipolar data, as Poisson encoding requires non-negative values. The positive (negative) channel retains only the positive (negative) values of the feature vector.
    \item \textbf{Min-Max Rescaling:} Each channel is transformed to the range $[0, 1]$. This sets the average spiking rate near $0.5$ spikes per timestep, balancing network excitation and energy efficiency. To increase sparsity, the upper bound can be lowered such that the range becomes $[0, x]$ where $x < 1$.
\end{enumerate}

The proposed method is straightforward to implement and eliminates the need for computationally expensive hyperparameter searches to find optimal values of $\mu_{\text{norm}}$ and $\sigma_{\text{norm}}$. Additionally it gracefully handles outliers while effectively setting the average spiking rate to an appropriate value.
 
\subsubsection{Encoding}
\label{subsec:encoding} 
The final preprocessing step uses Poisson encoding to generate spike trains from the normalised time series data. For a feature vector with values $x_i$, a spike train of length $T$ is generated for all $i=1, 2, ..., N$, following the pseudo-code in Algorithm~\ref{alg:poisson_generation}. This is repeated for all $K$ features, hence the resulting dataset contains $K$ features, with $N$ timestamps and $T$ timesteps per spike train. In total, there are $N \cdot K$ spike trains generated.  
For the results presented here, $T$ was chosen to be 20, balancing both encoding precision and computational costs.

\begin{algorithm}[!htb]
\caption{Generate Poisson Spike Train}
\label{alg:poisson_generation}
\begin{algorithmic}
\REQUIRE $T$ (number of timesteps)
\REQUIRE $x_i$ (firing rate)
\\
\STATE $spikeTrain \leftarrow \text{zeros}(T)$  \COMMENT{Initialize spike train}
\\
\FOR{$t = 1$ to $T$}
    \STATE $p \leftarrow x_i$ \COMMENT{Spike probability in bin}
    \STATE $u \sim \text{Uniform}(0,1)$ \COMMENT{Random number}
    \IF{$u < p$}
        \STATE $spikeTrain[t] \leftarrow 1$ \COMMENT{Generate spike}
    \ENDIF
\ENDFOR
\\
\RETURN $spikeTrain$
\end{algorithmic}
\end{algorithm}

\subsection{Spike Definitions and Predictive Metrics}
\label{subsec:models_metrics}
The concept of ``spikes'' in this research carries specific technical meaning within the context of unsupervised price-spike learning for financial time series prediction. These definitions provide the framework for evaluating the neural network's ability to identify significant market events.

\subsubsection{Real vs Fake Price Spikes}
A primary distinction in the proposed evaluation framework is between \emph{real} and \emph{fake} price spikes, which determines the fundamental accuracy of the SNN predictions. 

A real price spike occurs when the SNN emits a signal that precedes a significant price movement. This significance is quantified by comparing the subsequent price movement to a predefined threshold. Formally, we define the absolute percentage return at time $t$ as:
\begin{equation}
r_t = \left|\frac{X_{t+1}}{X_t} - 1\right|, \quad t = 1, 2, \ldots, n-1
\end{equation}
where $X_t$ represents the volume-weighted average price (VWAP) at time $t$. The threshold for determining significance is defined as the median of the intra-day absolute return series:
\begin{equation}
r_{\text{thresh}} = \text{median}(r_t), \quad t = 1, 2, \ldots, n-1
\end{equation}
For each price spike signal emitted by the network at time $t$, its strength is calculated over a subsequent window, $w$, of time:
\begin{equation}
S_\text{strength} = \frac{|r_{t+1}| + |r_{t+2}| + ... + |r_{t+w}|}{w}
\end{equation}
A price spike is classified as real if $S_{\text{strength}} > r_{\text{thresh}}$, indicating that the network has identified a price movement of above-median magnitude. All other spikes not meeting this criteria are considered fake spikes.

\subsubsection{Momentum vs Reversion Spikes}
Another categorisation of price spikes is based on the directional relationship between the preceding price trend and the subsequent movement. A \emph{momentum spike} occurs when a price movement continues in the same direction as the immediate trend preceding the current price. A \emph{reversion spike} occurs when a real spike is followed by a price movement in the opposite direction to the immediate preceding trend. Mathematically, a real price spike at time $t$, is subsequently classified as:
\begin{equation}
\text{Class} = 
\begin{cases}
\text{Momentum}, & \text{if } (X_t - X_{t-w}) \cdot (X_{t+w} - X_t) > 0 \\
\text{Reversion}, & \text{if } (X_t - X_{t-w}) \cdot (X_{t+w} - X_t) < 0 \\
\text{None}, & \text{otherwise}
\end{cases}
\end{equation}
where $w$ represents the window size for trend determination.

\subsubsection{Predictive Metrics}
\label{sssec:predictive metrics}
A range of predictive metrics are defined to support quantitative evaluation of performance. They are summarised in Table~\ref{tab:performance_metrics}

\begin{table}[!htb]
    \centering
    \caption{Performance Metrics for Evaluating SNN Models}
    \label{tab:performance_metrics}
    \footnotesize 
    \setlength{\tabcolsep}{2pt} 
    \begin{tabular}{|p{2.2cm}|c|p{3.5cm}|} 
        \hline
        \textbf{Metric} & \textbf{Equation} & \textbf{Purpose} \\
        \hline
        Spike Accuracy & $ \frac{N_{\text{predicted real}}}{N_{\text{predicted}}}$ & Measures the precision of identified signals \\
        \hline
        Momentum Spike Percentage & $ \frac{N_{\text{predicted momentum}}}{N_{\text{predicted}}}$ & Assesses suitability for momentum strategies \\
        \hline
        Spiking Rate & $ \frac{N_{\text{predicted}}}{N_{\text{total timestamps}}}$ & Quantifies model activation frequency \\
        \hline
        Real Spiking Rate & $ \frac{N_{\text{real spikes}}}{N_{\text{total timestamps}}}$ & Establishes baseline for comparison \\
        \hline
        True Positive Rate & $ \frac{N_{\text{predicted real}}}{N_{\text{real}}}$ & Measures sensitivity/recall \\
        \hline
        False Positive Rate & $ \frac{N_{\text{predicted fake}}}{N_{\text{fake}}}$ & Quantifies false signal generation \\
        \hline
    \end{tabular}
    
\end{table}

\subsubsection{The Penalised Spike Accuracy (PSA) Objective}
The selection of an appropriate optimisation objective is paramount for effective hyperparameter tuning with Bayesian optimisation. While price spike accuracy provides a natural target, it presents limitations when used in isolation. The primary limitation is that it results in a preference for lower SNN spiking rates and can give drastically different results for different time frames. To compel BO to discover models with both high accuracy and an \textit{actionable} spike rates, this work introduces the novel Penalised Spike Accuracy (PSA) metric:
\begin{equation}
\text{PSA} = \text{Spike Accuracy} \times \text{Penalty Factor}
\label{eq:PSA}
\end{equation}
The Penalty Factor exponentially diminishes the score if the model's output rate deviates too far from the empirically determined {Real Spike Rate} (RSR), thus rewarding balance:
\begin{equation}
\text{Penalty Factor} = \exp\left(-\max(|\text{SRD}| - \alpha, 0)\right)
\end{equation}
where $\alpha=0.05$ is a tolerance threshold and Spike Rate Deviation (SRD), is defined as the proportional difference from the true frequency (Eq~\ref{eq:SRD}).
\begin{equation}
\text{SRD} = \frac{\text{Spike Rate}}{\text{Real Spike Rate}} - 1
\label{eq:SRD}
\end{equation}

\subsection{Model Architectures}
Three SNN model architectures are proposed, sharing fundamental characteristics that form the foundation of this research. All implementations employ LIF neurons with preset membrane thresholds determined as a hyperparameter, and employ current-based synapses throughout the architecture. A standard simulation process describing the flow of data from input to output is constant across models, containing: input, propagation, and output. \textbf{Input} For each timestep $t=1, 2, ..., T$, $K$ spike trains are input to the network, where $T$ is the number of timesteps and $K$ is the number of features. \textbf{Propagation} Once the input spikes enter the network, they propagate according to the dynamics of LIF neurons.
The membrane potential, $V$, of an LIF neuron evolves according to the differential equation \cite{Gerstner1996}:
\begin{equation}
\beta \frac{dV}{dt} = -V + R I(t)
\label{eqw:LIF}
\end{equation}
Where $\beta$ is the membrane time constant (decay rate), $R$ is the membrane resistance and $I(t)$ is the input current (sum of weighted incoming spikes).
Upon receiving a spike, a neuron’s membrane potential is incremented by the synapse’s weight. If this potential exceeds the preset threshold, $V_{\text{thresh}}$, the neuron emits a spike that propagates forward. In the absence of a spike, the potential decays over time, and after a spike, the neuron’s potential resets by subtracting the membrane threshold from the current potential. Lastly, the neuron enters a brief refractory period of precisely one timestep to prevent immediate re-firing. \textbf{Output} In the final layer, a single spike train is outputted. The number of spikes in each spike train corresponds to the confidence that a price spike has occurred. If this number exceeds a decoding threshold, $D_{thresh}$, this is interpreted as the model predicting a price spike.
\subsection*{Model 1 - Double Input SNN}
\label{subsec:model1_gao}

\begin{figure}[t!]
    \centering
    
    \includegraphics[width=0.4\textwidth]{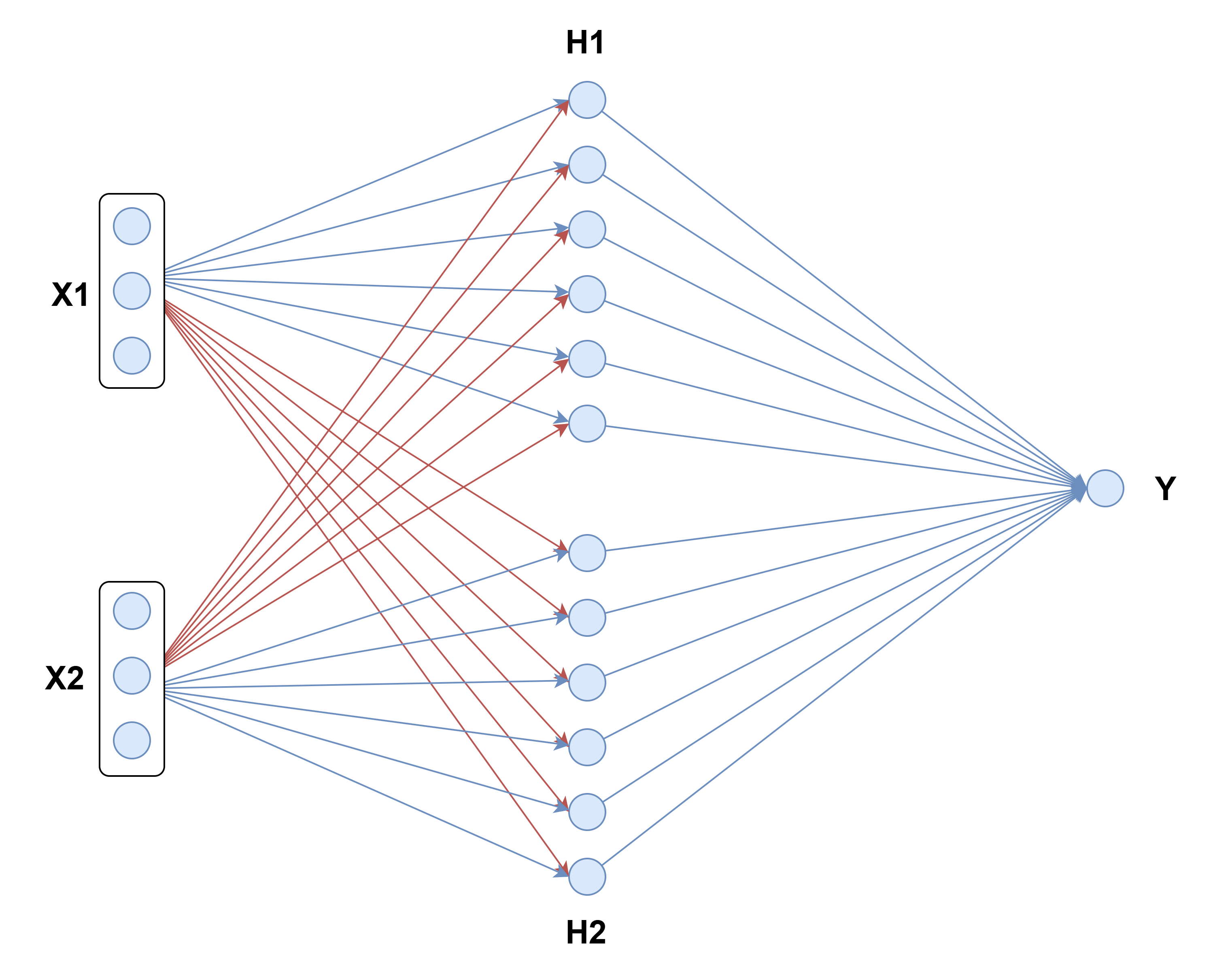}
    \caption{The extended architecture incorporating multiple time lags (additional nodes in $X_1 \text{ and }X_2$) and explicit inhibitory connections (in red).}
    \label{fig:double_input_comparison}
\end{figure}
The first model, an existing architecture proposed by Gao et~al. (2021) \cite{Gao2023}, is designed for the competitive temporal dynamics of STDP learning for price spike detection. Its structure (blue connections in Figure~\ref{fig:double_input_comparison}) is distinct from a standard feedforward network, instead opting for the creation of two distinct pathways. The input layer consists of two neurons, one in $X_1$ and one in $X_2$, processing the positive and negative price difference series at time $t$, respectively,  which are encoded into spike trains via Poisson rate encoding. The hidden layer is segregated into two sub-layers, $H_1$ and $H_2$, each exclusively connected to one input stream ($X_1 \to H_1$, $X_2 \to H_2$), ensuring structural separation. The single output LIF neuron receives full connectivity from $H_1$ and $H_2$, with its spikes serving as price-spike predictions.

Training employs the STDP rule, adjusting synaptic weights based on the temporal relationship between pre- and postsynaptic spikes, defined by an exponential window function \eqref{eq:stdp}. The network maintains stability through a synaptic homeostasis mechanism: weights are bounded between 0 and 1, and a 5\% reduction is applied to all layer weights if the mean weight exceeds 0.5. This process prevents unconstrained growth and maintains stable dynamics. The training process operates in an unsupervised manner, relying solely on historical data for real-time deployment viability. The hyperparameters, found in Table~\ref{tab:combined_hyperparameters_final} include the neuron, structural and synaptic plasticity parameters that were tuned using Bayesian optimisation.

\begin{table}[!tb]
    \centering
    \caption{Hyperparameters for Model 1 and Model 2}
    \label{tab:combined_hyperparameters_final}
    \footnotesize
    \setlength{\tabcolsep}{1.5pt} 
    \begin{tabular}{|l|l|p{4cm}|} 
    \hline
    \textbf{Parameter} & \textbf{Symbol} & \textbf{Function} \\
    \hline
    \multicolumn{3}{|l|}{\textit{\textbf{Neuron and Structural Parameters}}} \\
    \hline
    Decay Rate & $\beta$ & Rate of membrane potential decay \\
    Membrane Threshold & $V_{\text{thresh}}$ & Spike generation threshold \\ 
    Hidden Layer Neurons & $N_{hidden}$ & Size of the network's hidden layer \\
    Input Neurons & $N_{input}$ & Number of lags used as features \\
    \hline
    \multicolumn{3}{|l|}{\textit{\textbf{Synaptic Plasticity Parameters (Excitatory/Inhibitory)}}} \\
    \hline
    Potentiation Learning Rate  & $A_+ / B_+$ & Controls weight strengthening \\
    Depression Learning Rate  & $A_-/ B_-$ & Controls weight weakening \\
    Potentiation Time Constant  & $\tau_+/ \theta_+$ & Temporal window for  potentiation \\
    Depression Time Constant  & $\tau_-/ \theta_-$ & Temporal window for depression \\
    \hline
    \end{tabular}
\end{table}

\subsection*{Model 2: Double Input SNN (Extended)}
\label{subsec:model2_extended}
This extended architecture builds on Model 1 by incorporating additional temporal information and explicit inhibitory connections (Figure~\ref{fig:double_input_comparison}), to enhance predictive performance. The model maintains the core segregated structure but introduces two significant enhancements. First, the input layer is expanded to incorporate $k$ multiple time lags for price differences, providing the network with comprehensive temporal information across different time scales. Input encoding is defined by $X_{1i} \leftarrow (P_n - P_{n-i})_+$ and $X_{2i} \leftarrow (P_n - P_{n-i})_-$, for $i \in [1, k]$. Second, {explicit inhibitory connections} are integrated from $X_1$ to $H_2$ and from $X_2$ to $H_1$. These inhibitory synapses, constrained to weights between 0 and -1, introduce direct competition, enhancing the model's ability to isolate price momentum by mutually suppressing activity during noisy, bidirectional fluctuations. Essentially, the network architecture is optimised to isolate moments of price momentum over $k$ time lags, interpreted as price spikes that constitute actionable trading signals.

The training methodology maintains the core unsupervised STDP framework of Model 1, but critically modifies weight adjustments for inhibitory synapses. For inhibitory connections, weight updates $\Delta w$ are subtracted from existing weights, ensuring their inhibitory nature. The specific adjustment rule is: $\Delta w = - W(\Delta t)$, where $W(\Delta t)$ follows the standard exponential window STDP function (Equation~\ref{eq:stdp}), ensuring stable learning dynamics while improving selectivity.

Model 2 includes all hyperparameters from Model 1, plus additional parameters for governing the inhibitory STDP dynamics and the increased complexity of the input layer (Table~\ref{tab:combined_hyperparameters_final}), which were determined using Bayesian optimisation.

\subsection*{Model 3: Fully Connected SNN}
\label{subsec:model3_fully_connected}
The final model serves as a supervised learning reference to benchmark against the unsupervised Models 1 and 2. It employs a conventional, fully connected feedforward Spiking Neural Network (SNN) architecture with two hidden layers and two output neurons. This structure processes a broad range of financial features, including Returns, Volatility, and Volume, encoded into spike trains using Poisson rate encoding. Unlike the STDP-trained models, the architecture allows all input features to influence all subsequent neurons. The two output neurons enable binary classification for identifying significant price movements (real spikes) versus non-significant movements, allowing for direct optimisation against known labels.

Training is performed using supervised learning via Spike-based Backpropagation Through Time (BPTT). The loss function is a mean squared error count (target rates: $80\%$ of the time for the correct class, $20\%$ otherwise), encouraging targeted firing. Optimisation uses the Adam algorithm \cite{Eshraghian2023, kingma2017adammethodstochasticoptimization} with the fast sigmoid as the surrogate gradient descent method.

The following fixed hyperparameters were manually selected to balance capacity, stability, and efficiency: the learning rate was set to $0.005$; the number of hidden neurons ($N_{\text{hidden}}$) was $128$; the membrane threshold ($V_{\text{thresh}}$) was $1$; and temporal dependencies were captured using time lags of $1$, $3$, and $5$. The input features included Returns, Volatility, and Volume, selected for their strong temporal dependencies.


\subsection{Experimental Design}
\label{sec:experimental_design}
Two complementary experimental frameworks were designed to rigorously evaluate the proposed methods and models: a rolling train-test experiment to assess temporal predictive performance, and a hyperparameter optimisation experiment to validate the novel Penalised Spike Accuracy (PSA) metric.

\subsubsection*{Experiment 1: Comparison of Objective Values}
\label{subsec:hyperparameter_optimisation_experiment}
To assess the effectiveness of PSA against standard Spike Accuracy (SA), we employ Bayesian optimisation over 100 iterations for both metrics. In each iteration, models are trained and evaluated using sequential batches of $5000$ timestamps to enhance generalisability and mitigate overfitting. This process is repeated for both Model 1 and Model 2. The hyperparameters yielding the highest objective score for each metric (SA and PSA) are selected as optimal for subsequent analysis. These optimal parameters are used to train the definitive models for final comparison. Further analysis isolates the most impactful hyperparameters and maps their search spaces, detailed in Table~\ref{tab:param_search_space}.

\begin{table}[!tb]
\centering
\caption{Hyperparameter Search Space}
\label{tab:param_search_space}
\begin{tabular}{|l|l|l|}
\hline
\textbf{Parameter} & \textbf{Values} & \textbf{Scale} \\
\hline
$A_{+}$ & [0.0001, 0.01] & Logarithmic \\
$A_{-}$ & [$A_{+}$ - 0.001, $A_{+}$] & Logarithmic \\
$\tau_{+}$ & [5, 100] & Linear (Integer) \\
$\tau_{-}$ & [$\tau_{+}$ - 5, $\tau_{+}$ + 5] & Linear (Integer) \\
$B_{+}$ & [0.0001, 0.01] & Logarithmic \\
$B_{-}$ & [$B_{+}$ - 0.001, $B_{+}]$ & Logarithmic \\
$\theta_{+}$ & [5, 100] & Linear (Integer) \\
$\theta_{-}$ &[$\theta_{+}$ - 5, $\theta_{+}$ + 5] & Linear (Integer) \\
$\beta$ &[ 0.5, 0.99]  & Linear (step=0.01)\\
$V_{thresh}$ & [0.8, 2.5]  & Linear (step=0.1)\\
$D_{thresh}$ & [4, 16] & Linear (Integer) \\
$N_{input}$ & [1, 10] & Linear (Integer) \\
$N_{hidden}$ & \{16, 32, 64, 128\} & Categorical \\
\hline
\end{tabular}

\end{table}

\subsubsection*{Experiment 2: Comparison of Model Performance}
\label{subsec:rolling_train_test}
The second experiment compares the predictive performance of all models using a day-by-day rolling window structure to replicate real-world data workflows. The dataset consists of high-frequency intra-day price series spanning $19$ consecutive trading days in February 2015. The model is trained on one day's data and tested on the subsequent day. This process rolls forward sequentially: Day $i$ trains for prediction on Day $i+1$. This methodology ensures that each day (except the first and last) serves exactly once as a training set and once as a testing set. This design prevents future information from contaminating predictions, maintaining the integrity and validity of the results. The key performance metrics for evaluation are summarised in Table~\ref{tab:performance_metrics}.

\subsection{SNN Based Momentum Strategy}
The SNN based momentum strategy combines the SNN's ability to detect significant temporal patterns (spikes) with established momentum-trading logic. 
The core principle of momentum trading suggests that major price movements typically continue in their current direction due to market inertia. This manifests as autocorrelation in price movements, where recent price changes predict short-term future movements. 
When the SNN emits a spike signal at time $t$, a position is initiated. The trade's direction is determined by analysing the immediate preceding price trend using a look-back window of $n$ timestamps (default 3), quantified by a position flag, $F_t$ (Equation~\ref{eq:position_flag}). 
\begin{equation}
F_t = P_t - \frac{1}{n}\sum_{i=1}^{n} P_{t-i}
\label{eq:position_flag}
\end{equation}
A positive flag (current price below recent average) suggests a downward trend, triggering a short position; a negative flag suggests an upward trend, prompting a long position (Algorithm~\ref{alg:momentum_strategy}). This ensures the neural network identifies the opportune entry time, while the momentum calculation dictates the trade's direction.
\begin{algorithm}[!h]
\caption{Spike-Based Momentum Strategy}
\label{alg:momentum_strategy}
\begin{algorithmic}[1]
\IF{\texttt{$F_t$} $<$ 0}
    \STATE Enter short position
\ELSIF{\texttt{$F_t$} $>$ 0}
    \STATE Enter long position
\ENDIF
\end{algorithmic}
\end{algorithm}
The back-testing framework integrates the spike signal for entry timing. Upon spike detection, the trade is executed as a market order at the VWAP of the subsequent timestamp. Key back-testing parameters ensure consistent evaluation: initial capital is normalised to 1 unit, trades use full capital (1$\times$ leverage), and only one position is open at any time, maintained for a predetermined duration before closing at the exit timestamp's VWAP.

Performance is evaluated using metrics detailed in Table~\ref{tab:trading-label}, which assess both profitability and risk characteristics. Results are benchmarked against random and naive momentum strategies to quantify the value added by the SNN approach.

\begin{table}[!tb]
    \centering
    \caption{Trading Performance Metrics and Importance}
    \label{tab:trading-label}
    \footnotesize
    \begin{tabular}{|l|p{0.6\columnwidth}|}
        \hline
        \textbf{Metric} & \textbf{Importance} \\
        \hline
        Cumulative Return & Measures overall strategy effectiveness regardless of timeframe \\
        Sharpe Ratio & Evaluates return quality by accounting for volatility and risk taken \\
        Maximum Drawdown & Quantifies worst-case capital decline, essential for risk management \\
        Win Rate & Indicates strategy consistency and psychological sustainability \\
        Profit Factor & Assesses overall profitability efficiency and capital utilisation \\
        Profit-Loss Ratio & Reveals if winning trades sufficiently outweigh losing trades \\
        Expectancy & Provides mathematical expectation of profit per trade over time \\
        \hline
    \end{tabular}
        
\end{table}

\section{Results}

\subsection{Hyperparameter Optimisation}
\label{sec:hyperparameter_analysis}

Following Experiment 1, Bayesian optimisation was conducted for 100 iterations on both Model 1 and Model 2 across two objective metrics: Spike Accuracy (SA) and Predictive Spike Accuracy (PSA). The results, gathered using the Optuna library \cite{akiba2019optuna}, are presented in Table~\ref{tab:model-hyperparam-results}.

\begin{table}[!htb]
\centering
\caption{Hyperparameters for Models}
\label{tab:model-hyperparam-results}

\begin{tabular}{|l|c|c|c|c|}
\hline
\multirow{2}{*}{\textbf{Parameter}} & \multicolumn{2}{c|}{\textbf{Model 1}} & \multicolumn{2}{c|}{\textbf{Model 2}} \\
\cline{2-5}
& \textbf{SA} & \textbf{PSA} & \textbf{SA} & \textbf{PSA} \\
\hline
$A_{+}$ & 0.0037 & 0.0067 & 0.0096 & 0.0012 \\
$A_{-}$ & 0.0032 & 0.0063 & 0.0093 & 0.0009 \\
$B_{+}$ & -- & -- & 0.0059 & 0.0016 \\
$B_{-}$ & -- & -- & 0.0052 & 0.0009 \\
$\tau_{+}$ & 45 & 71 & 74 & 54 \\
$\tau_{-}$ & 42 & 72 & 72 & 58 \\
$\theta_{+}$ & -- & -- & 44 & 51 \\
$\theta_{-}$ & -- & -- & 44 & 51 \\
$\beta$ & 0.96 & 0.79 & 0.87 & 0.86 \\
$ V_{\text{thresh}}$  & 2.20 & 0.80 & 2.50 & 2.00 \\
$N_{\text{hidden}}$  & 16 & 32 & 128 & 64 \\
$N_{\text{input}} $  & -- & -- & 1 & 3 \\
$D_{\text{thresh}}$  & 12 & 9 & 12 & 11 \\
\hline
Objective Value & 0.90 & 0.76 & 0.77 &0.71\\
Spike Rate Deviation & -0.69 & -0.01 & -0.43 & 0.078\\
\hline
\end{tabular}

\end{table}

\begin{figure*}[!tb] 
    \centering
    \setlength{\tabcolsep}{0.4cm} 
    \begin{tabular}{cc} 
        \hspace*{-0.75cm}
        \includegraphics[width=0.95\columnwidth]{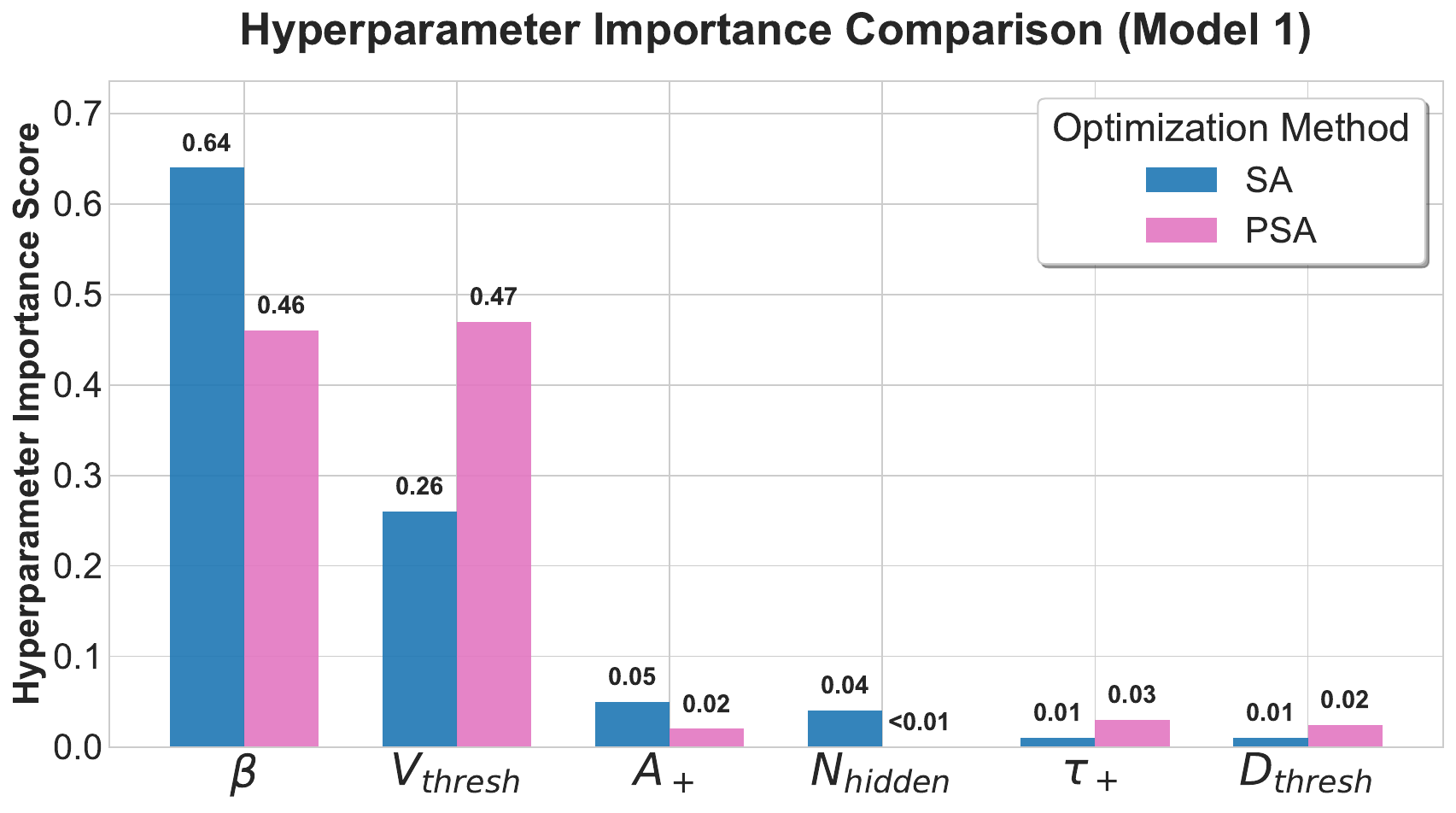} & 
        \includegraphics[width=0.95\columnwidth]{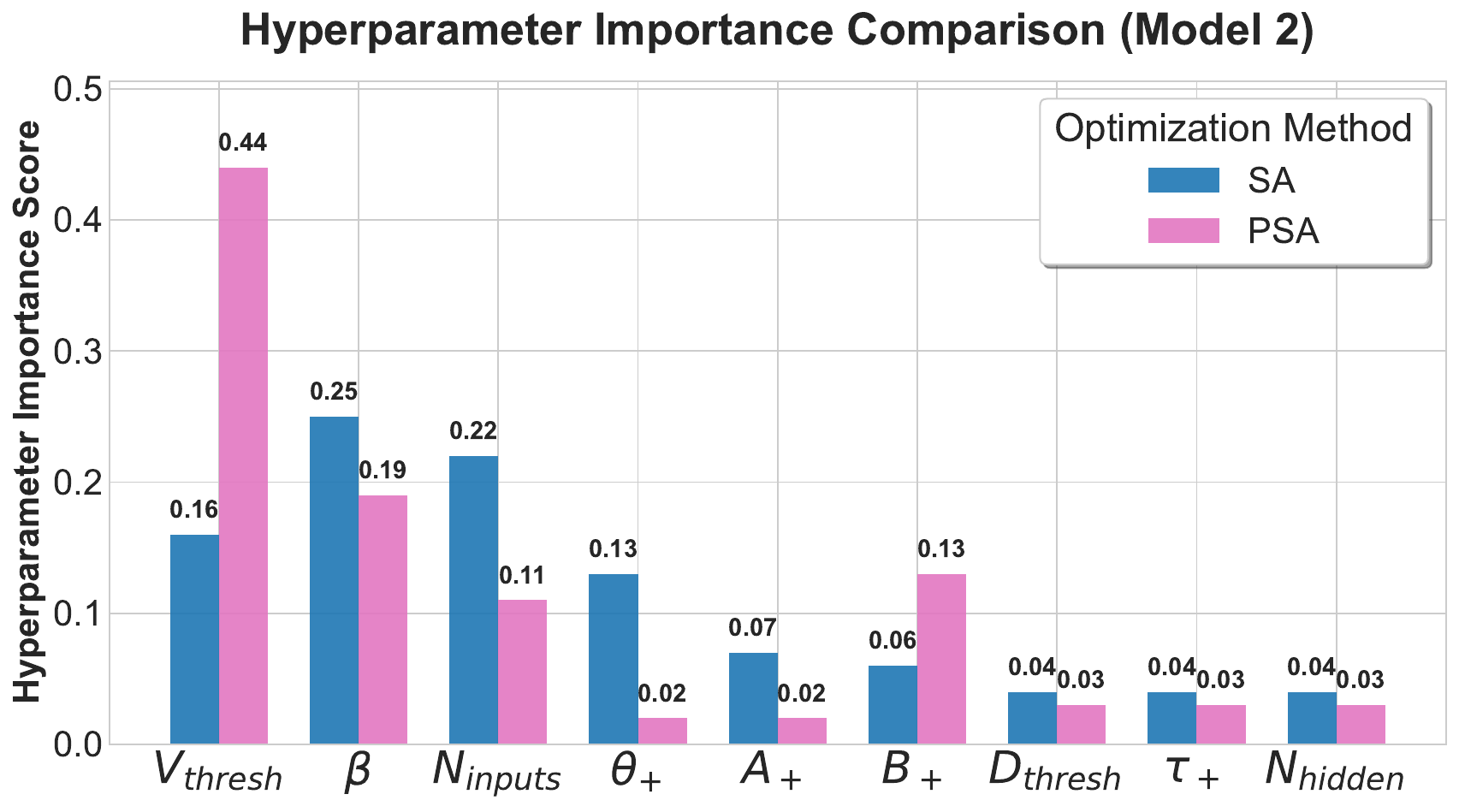} \\
        
        \footnotesize (a) Model 1 & 
        \footnotesize (b) Model 2
    \end{tabular}

    \caption{Parameter importance ranking for Models~1~\&~2, explored for performance metrics SA and PSA.}
    \label{fig:param_importances}
\end{figure*}

\subsubsection{Analysis of Optimal Parameters}
The optimised hyperparameters in Table~\ref{tab:model-hyperparam-results} reveal key differences dictated by the SA and PSA objectives. The table lists the highest objective score achieved and the corresponding spike rate deviation.

For Model 1, PSA generally selected larger excitatory learning rates ($A_{+}, A_{-}$) and learning windows ($\tau_{+}, \tau_{-}$) compared to SA. This, combined with a low membrane threshold ($V_{\text{thresh}} = 0.80$) and spike decoding threshold ($D_{\text{thresh}} = 9$), promotes higher network activity and increased spike rates. Conversely, the SA parameters (e.g., $V_{\text{thresh}} = 2.2$) favor conservative network action and higher selectivity, resulting in much lower spike rates.

Model 2 exhibited different trends, underscoring the challenge in tuning STDP-trained networks. PSA selected low learning rates and time constants for both excitatory and inhibitory synapses, alongside a higher membrane threshold ($V_{\text{thresh}}=2.0$). This suppression is likely a necessary control mechanism, as Model 2 PSA selected three input features ($N_{\text{input}}=3$), providing a considerable volume of input spiking activity. Model 2 SA, however, selected only one input feature ($N_{\text{input}}=1$) and could thus afford larger learning rates. SA also selected the upper bound membrane threshold ($V_{\text{thresh}}=2.5$), confirming its preference for conservative behavior.

Across both models, $V_{\text{thresh}}$, $\beta$, and $D_{\text{thresh}}$ were consistently lower for PSA than SA, further supporting the observation that PSA optimization drives higher network activity. Observing the Spike Rate Deviation (Table~\ref{tab:model-hyperparam-results}) confirms that SA consistently yields large negative deviations (lower spike rates), while PSA results in minor fluctuations close to the true rate. The higher objective values for Model 1 relative to Model 2 suggest greater training difficulty for the latter. Overall, while PSA shows better performance regarding adherence to the true spike rate, further analysis is required to determine its overall practical efficacy.

\subsubsection{Parameter Importance}
\label{subsec:parameter_importance}

As shown in Figure~\ref{fig:param_importances}, the membrane threshold ($V_{\text{thresh}}$) and the decay rate ($\beta$) were consistently the two most critical hyperparameters across both models and metrics. Importance was highly skewed toward these two parameters for Model 1. In contrast, Model 2 showed a more even distribution of importance, likely due to its added complexity (inhibitory synapses) requiring tuning of multiple parameters for improved results. For both models, PSA increased the skew toward the membrane threshold, suggesting that fine control over spiking rates was primarily achieved by adjusting $V_{\text{thresh}}$ to suppress or increase spiking activity. Other parameters, such as the depression learning rates and windows, had minimal effect, as their optimal values were generally dependent on their potentiating counterparts.

\subsection{Predictive Performance}
\label{sec:train-test_experiment}
Table~\ref{tab:performance_comparison} summarises key metrics for models optimised via SA and PSA respectively, alongside Model 3. The random model had a 50\% chance of spiking at any timestamp and adds additional context to these metrics. These measures provide insight into each optimisation method’s ability to balance accuracy against network activity and error rates.
\begin{table}[!b]
    \centering
    \caption{Comparison of the predictive performance of all models and a random baseline}
    \label{tab:performance_comparison}
    \resizebox{\columnwidth}{!}{%
        \scriptsize 
        \begin{tabular}{p{1.9cm} c c c c c c} 
            \toprule
            \textbf{Metric} & \textbf{Random} & \multicolumn{2}{c}{\textbf{Model 1}} & \multicolumn{2}{c}{\textbf{Model 2}} & \textbf{Model 3} \\
            \cmidrule(lr){3-4} \cmidrule(lr){5-6}
            & & \textbf{SA} & \textbf{PSA} & \textbf{SA} & \textbf{PSA} & \\
            \midrule
            Spike Acc. ($\%$) & $59.85$ & $73.85$ & $62.67$ & $67.13$ & $65.11$ & $69.35$\\
            Mom. Spike ($\%$) & $54.47$ & $54.34$ & $54.49$ & $55.08$ & $55.06$ & $55.09$\\
            TPR ($\%$) & $50.00$ & $12.96$ & $67.33$ & $12.51$ & $48.62$ & $36.88$\\
            FPR ($\%$) & $50.00$ & $7.03$ & $59.81$ & $9.13$ & $38.83$ & $24.30$\\
            Spiking Rate ($\%$) & $50.00$ & $10.58$ & $64.31$ & $11.15$ & $44.69$ & $31.83$\\
            \bottomrule
        \end{tabular}%
    }

\end{table}

Models optimised with PSA exhibit significantly higher spike rates and True Positive Rates (TPR), suggesting that this method drives greater network sensitivity and spiking activity. Conversely, SA optimisation leads to high precision (Spike Acc.) but much lower spiking rates and TPR, indicating a highly conservative firing strategy that may improve efficiency but risks missing true events. All models demonstrate a consistent Momentum Spike Percentage near the baseline, confirming reliable directional prediction regardless of the specific optimisation objective. All models successfully surpass the random baseline in Spike Accuracy, validating the utility of the SNN approach. Model 3, representing a supervised learning baseline, shows performance that is generally average compared to the diverse outcomes of the unsupervised SA and PSA methods.

\subsection{Trading Performance}
\label{sec:trading_performance}

Trading performance is assessed and compared against two baselines: a naive momentum strategy (following Algorithm~\ref{alg:momentum_strategy} at every step) and a random trading approach (50\% spike chance). Each model was run three times (100 for the random model), and results were averaged. For a fair comparison, cumulative returns in Table~\ref{tab:trading_performance_scaled} are scaled to $1,000$ trades per day, totalling $19,000$ trades over the one-month period. All other metrics are reported using their original values.

\begin{table}[!tb]
    \caption{Trading Performance Comparison Across Models (Returns Scaled to $1,000$ Trades per Day)}
    \label{tab:trading_performance_scaled}
    \resizebox{\columnwidth}{!}{%
        \scriptsize 
        \begin{tabular}{p{1.9cm} c c c c c c c} 
            \toprule
            \textbf{Metric} & \multicolumn{2}{c}{\textbf{Model 1}} & \multicolumn{2}{c}{\textbf{Model 2}} & \textbf{Model 3} & \textbf{Naive} & \textbf{Random} \\
            \cmidrule(lr){2-3} \cmidrule(lr){4-5}
            & \textbf{SA} & \textbf{PSA} & \textbf{SA} & \textbf{PSA} & & & \\
            \midrule
            Cum. Return ($\%$) & $15.49$ & $15.48$ & $13.63$ & $\mathbf{17.44}$ & $12.44$ & $13.49$ & $12.71$ \\
            Sharpe Ratio & $10.70$ & $16.48$ & $16.72$ & $\mathbf{19.72}$ & $15.91$ & $16.32$ & $17.99$ \\
            Max. DD ($\%$) & $5.85$ & $2.52$ & $2.25$ & $2.69$ & $2.95$ & $\mathbf{1.76}$ & $2.77$ \\
            Win Rate ($\%$) & $52.63$ & $52.60$ & $52.89$ & $\mathbf{53.49}$ & $52.79$ & $52.42$ & $52.28$ \\
            Profit Factor & $1.15$ & $1.21$ & $1.16$ & $\mathbf{1.22}$ & $1.13$ & $1.18$ & $1.17$ \\
            Profit-Loss Ratio & $1.00$ & $\mathbf{1.09}$ & $1.03$ & $1.06$ & $1.01$ & $1.07$ & $1.07$ \\
            Expectancy {($\times 10^{-6}$)} & $8.15$ & $8.15$ & $7.17$ & $\mathbf{9.18}$ & $6.55$ & $7.10$ & $6.69$ \\
            \bottomrule
        \end{tabular}%
    } 

\end{table}

The models optimised with the PSA metric consistently outperform their SA counterparts and both baselines across most key performance measures (Table~\ref{tab:trading_performance_scaled}). Notably, Model 2–PSA achieves the highest cumulative return ($17.44 \%$), the strongest risk-adjusted return ($\text{Sharpe}=19.71$), the best win rate ($53.49 \%$), Profit Factor ($1.22$), and Expectancy ($9.18\times 10^{-6}$). Although its Max Drawdown ($2.69 \%$) is slightly higher than the Naive strategy ($1.76 \%$), its overall profile is superior.

Comparing optimisation methods directly, PSA significantly improved risk control for Model 1, halving the drawdown ($2.52 \%$ vs $5.85 \%$) and substantially boosting the Sharpe Ratio ($16.48$ vs $10.70$), despite a negligible change in headline return ($15.48 \%$ vs $15.49 \%$). Model 2 saw a more pronounced benefit from PSA, with return increasing from $13.63 \%$ to $17.44 \%$ and Sharpe Ratio enhancing from $16.72$ to $19.71$.

In contrast, Model 3 (supervised) yielded moderate outcomes, falling short of both PSA-optimised models in cumulative return ($12.44 \%$) and Sharpe Ratio ($15.91$). Its maximum drawdown ($2.95 \%$) was acceptable, but its Profit Factor ($1.13$) and Expectancy ($6.55\times 10^{-6}$) were lower than the STDP-trained models.

\begin{figure}[!tb]
    \centering
    \begin{tabular}{c} 
        \includegraphics[width=0.95\columnwidth]{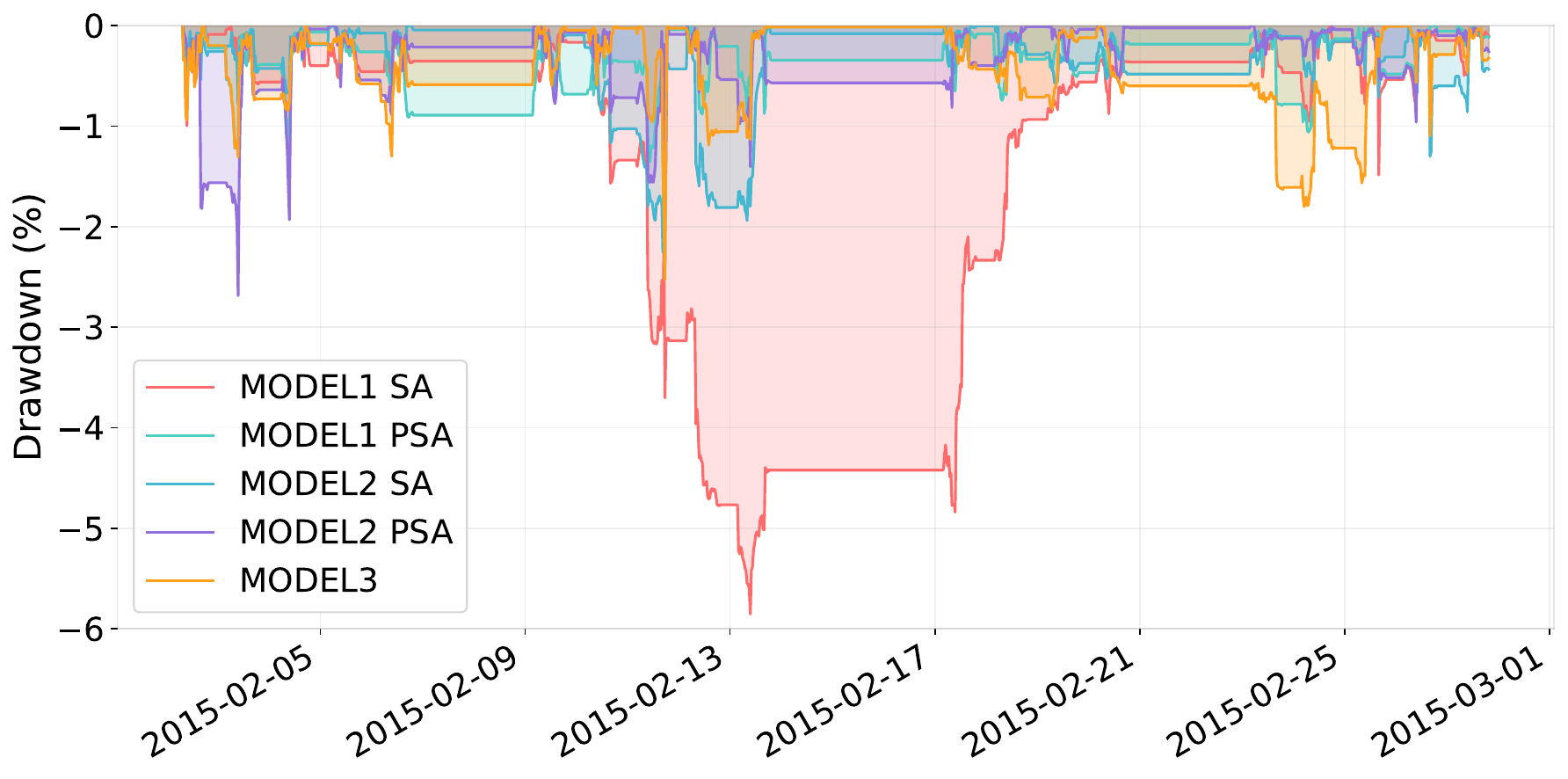} \\
        \footnotesize (a) Drawdown \\[0.3cm] 
        
        \includegraphics[width=0.95\columnwidth]{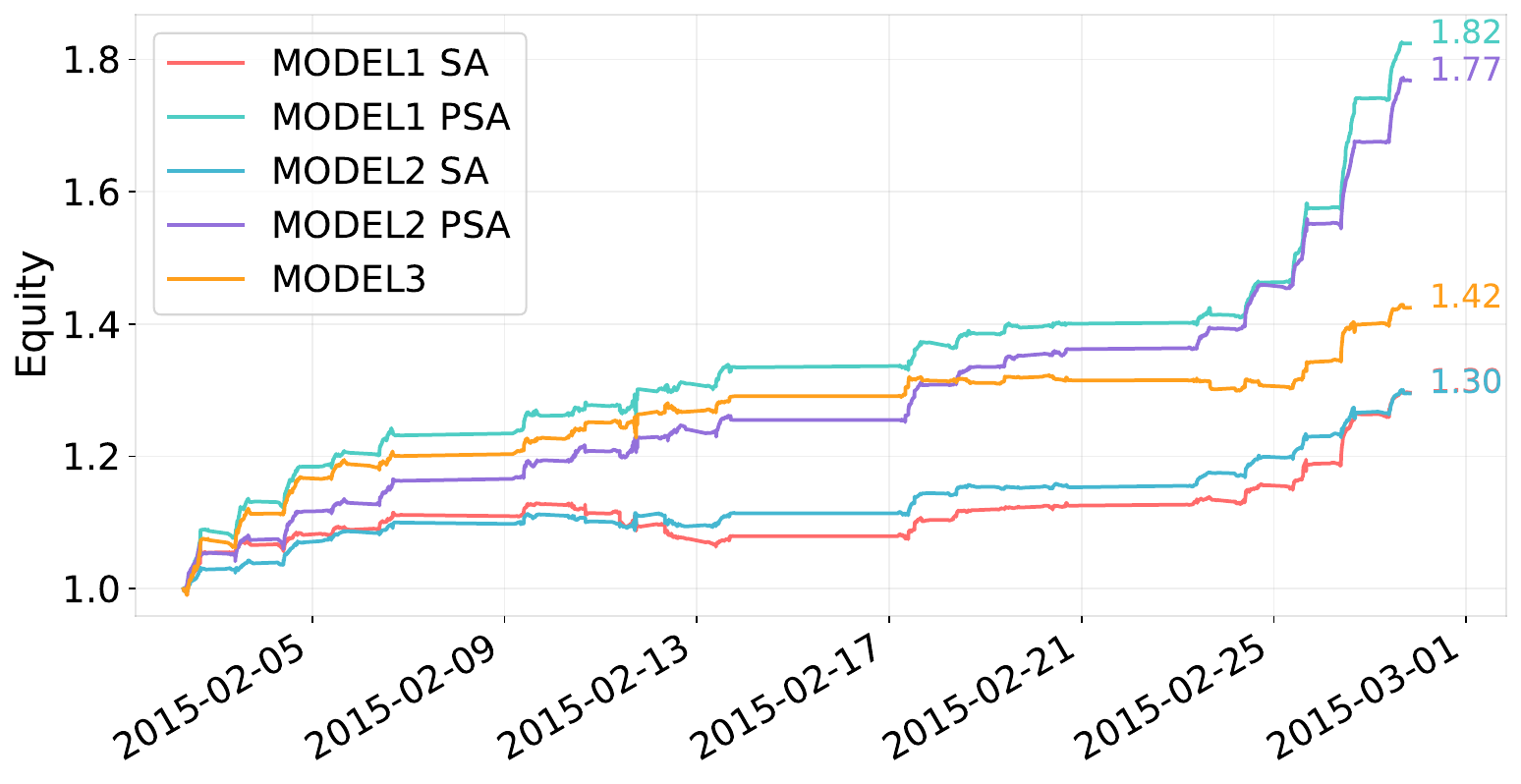} \\
        \footnotesize (b) Equity (unscaled)
    \end{tabular}
    
    \caption{Equity (unscaled) and Drawdown over time.}
    \label{fig:equity_drawdown_comparison}
\end{figure}

The equity and drawdown curves (Figure~\ref{fig:equity_drawdown_comparison}) further confirm the superiority of the PSA-optimised models. The drawdown plot highlights Model 1–SA's susceptibility to high volatility, particularly around the 12th and 13th day. In the unscaled equity curves, both PSA variants markedly outperform their SA counterparts, achieving total returns over $70 \%$ for the month, while SA models remain below $30 \%$. Model 3 achieved a moderate $42 \%$ return. The primary driver of the PSA models' higher returns is their higher spiking rate, enabling them to exploit more trading opportunities, as detailed by the total trade counts in Table~\ref{tab:model_performance}.
\begin{table}[!t]
    \centering
    \caption{Model Performance Comparison (unscaled)}
    \label{tab:model_performance}
    \begin{tabular}{lrrr}
        \hline
        \textbf{Model} & \textbf{Final Equity} & \textbf{Total Return} & \textbf{Num Trades} \\
        \hline
        Model 1-SA  & 1.296311 & 29.63\% & 36,336 \\
        Model 1-PSA & 1.824112 & 82.41\% & 101,126 \\
        Model 2-SA  & 1.295183 & 29.52\% & 41,141 \\
        Model 2-PSA & 1.768009 & 76.80\% & 83,673 \\
        Model 3     & 1.424655 & 42.47\% & 51,229 \\
        \hline
    \end{tabular}

\end{table}

\section{Conclusion}

Building upon the foundational work of Gao et al., this research successfully validated the application of Spike-Timing-Dependent Plasticity (STDP)-trained Spiking Neural Networks for high-frequency trading price-spike prediction using a reproducible framework to rigorously compare three architectures: the original replication (Model 1), a novel extended SNN featuring inhibitory synapses and enriched temporal inputs (Model 2), and a supervised feedforward SNN (Model 3).

Key achievements include the design of Model 2 and the effective application of Bayesian optimisation steered by a Penalised Spike Accuracy (PSA) objective. This metric reliably tuned the networks, ensuring close alignment with empirical spiking rates (e.g., Model 2 deviated by only $+0.07$). Crucially, backtesting a HFT momentum strategy demonstrated that all SNN models outperformed random benchmarks and decisively surpassed the supervised Model 3, with Model 2 (PSA) yielding the strongest trading performance, highest cumulative returns, and superior risk-adjusted metrics (Table \ref{tab:trading_performance_scaled}), thereby confirming the predictive power of the novel architecture combined with precise, PSA-driven hyperparameter tuning.

While this study establishes the efficacy of STDP-trained SNNs for HFT, several challenges remain. The current quantitative finance literature would benefit from additional peer-reviewed SNN benchmarks and standardization of SNN-based metrics for financial time series analysis (such as those presented in this work) to better facilitate comparison efforts and pave the way towards standardized deployment on neuromorphic hardware, allowing improved quantification of energy and latency performance essential for HFT applications.

\section{Funding}
This research was supported through the NimbleAI project, funded via the Horizon Europe Research and Innovation programme (Grant Agreement 101070679), and UKRI under the UK government’s Horizon Europe funding guarantee (Grant Agreement 10039070); the Horizon Europe AIDA4Edge project (Grant Agreement 101160293); and the EPSRC Edgy Organism project (EP/Y030133/1).

\bibliographystyle{IEEEtran}
\bibliography{refs} 

@ARTICLE{Gao2023,
  author={K. Gao and W. Luk and S. Weston},
  title={High-Frequency Trading and Financial Time-Series Prediction with Spiking Neural Networks},
  journal={Wilmott},
  year={2023},
}

@INPROCEEDINGS{Reid2013,
  author={D. Reid and A. J. Hussain and H. Tawfik},
  booktitle={The 2013 International Joint Conference on Neural Networks (IJCNN)}, 
  title={Spiking neural networks for financial data prediction}, 
  year={2013},
  pages={1--10},
  doi={10.1109/IJCNN.2013.6707140},
}

@PHDTHESIS{KozdonThesis2018,
  author={K. Kozdon},
  title={{STDP} for clustering spatio-temporal data},
  school={University College London},
  year={2018},
}

@ARTICLE{Yamazaki2022,
  author={K. Yamazaki and V.-K. Vo-Ho and D. Bulsara and N. Le},
  title={Spiking Neural Networks and Their Applications: A Review},
  journal={Brain Sciences},
  volume={12},
  number={7},
  pages={863},
  year={2022},
  doi={10.3390/brainsci12070863},
}

@ARTICLE{Eshraghian2023,
  author={Eshraghian, Jason K. and Ward, Max and Neftci, Emre O. and Wang, Xinxin and Lenz, Gregor and Dwivedi, Girish and Bennamoun, Mohammed and Jeong, Doo Seok and Lu, Wei D.},
  journal={Proceedings of the IEEE}, 
  title={Training Spiking Neural Networks Using Lessons From Deep Learning}, 
  year={2023},
  volume={111},
  number={9},
  pages={1016-1054},
  doi={10.1109/JPROC.2023.3308088}
}

@ARTICLE{Lobo2020,
  author={J. L. Lobo and J. Del Ser and A. Bifet and N. Kasabov},
  title={Spiking Neural Networks and online learning: An overview and perspectives},
  journal={Neural Networks},
  volume={121},
  pages={88--100},
  year={2020},
  doi={10.1016/j.neunet.2019.09.003},
}

@ARTICLE{Roy2019,
  author={K. Roy and A. Jaiswal and P. Panda},
  title={Towards spike-based machine intelligence with neuromorphic computing},
  journal={Nature},
  volume={575},
  number={7784},
  pages={607--617},
  year={2019},
  doi={10.1038/s41586-019-1603-9},
}

@ARTICLE{tavanaei2018deep,
  author={A. Tavanaei and M. Ghodrati and S. R. Kheradpisheh and T. Masquelier and A. Maida},
  title={Deep learning in spiking neural networks},
  journal={Neural Networks},
  volume={111},
  pages={47--63},
  year={2018},
  doi={10.1016/j.neunet.2018.12.002},
}

@INPROCEEDINGS{bergstra2013hyperparams,
  author={J. Bergstra and D. Yamins and D. D. Cox},
  title={Making a Science of Model Search: Hyperparameter Optimization in Hundreds of Dimensions for Vision Architectures},
  booktitle={Proceedings of the 30th International Conference on Machine Learning (ICML)},
  volume={28},
  year={2013},
}

@INPROCEEDINGS{akiba2019optuna,
  author={T. Akiba and S. Sano and T. Yanase and T. Ohta and M. Koyama},
  title={Optuna: A Next-Generation Hyperparameter Optimization Framework},
  booktitle={The 25th ACM SIGKDD International Conference on Knowledge Discovery \& Data Mining},
  pages={2623--2631},
  year={2019},
}

@ARTICLE{KimTaeYoon2004Annf,
  author={T. Y. Kim and K. J. Oh and C. Kim and J. D. Do},
  title={Artificial neural networks for non-stationary time series},
  journal={Neurocomputing (Amsterdam)},
  volume={61},
  pages={439--447},
  year={2004},
}

@ARTICLE{sezer2020financial,
  author={O. B. Sezer and M. U. Gudelek and A. M. Ozbayoglu},
  title={Financial time series forecasting with deep learning: A systematic literature review: 2005--2019},
  journal={Applied Soft Computing},
  volume={90},
  pages={106181},
  year={2020},
  doi={10.1016/j.asoc.2020.106181},
}

@INBOOK{Chakraborti,
  author={A. Chakraborti and M. Patriarca and M. S. Santhanam},
  title={Financial Time-series Analysis: a Brief Overview},
  booktitle={Econophysics of Markets and Business Networks},
  pages={51--67},
  year={2007},
  doi={10.1007/978-88-470-0665-2_4},
}

@INPROCEEDINGS{kohda2021hft,
  author={S. Kohda and K. Yoshida},
  title={Characteristics and Forecast of High-Frequency Trading},
  booktitle={Proceedings of the University of Tsukuba Research Presentations},
  year={2021},
}

@ARTICLE{davidson2021digital,
  author={S. Davidson and S. B. Furber},
  title={Comparison of Artificial and Spiking Neural Networks on Digital Hardware},
  journal={Frontiers in Neuroscience},
  volume={15},
  year={2021},
  doi={10.3389/fnins.2021.651141},
}

@MISC{frazier2018bayesian,
  author={P. I. Frazier},
  title={A Tutorial on {Bayesian} Optimization},
  year={2018},
  month={July},
  note={arXiv preprint arXiv:1807.02811},
}

@INPROCEEDINGS{seabold2010statsmodels,
  author={S. Seabold and J. Perktold},
  title={Statsmodels: Econometric and Statistical Modeling with Python},
  booktitle={Proceedings of the 9th Python in Science Conference},
  year={2010},
}

@BOOK{durbin2012timeseries,
  author={J. Durbin and S. J. Koopman},
  title={Time Series Analysis by State Space Methods},
  publisher={Oxford University Press},
  year={2012},
}

@INPROCEEDINGS{parsa2019bayesian,
  author={M. Parsa and R. M. Patton and J. P. Mitchell and T. E. Potok and C. D. Schuman and K. Roy},
  title={Bayesian-based Hyperparameter Optimization for Spiking Neuromorphic Systems},
  booktitle={2019 IEEE International Conference on Big Data (Big Data)},
  pages={4477--4484},
  year={2019},
  doi={10.1109/BigData47090.2019.9006273},
}

@article{Bi1998,
  author  = {Bi, Guo-Qiang and Poo, Mu-ming},
  title   = {Synaptic modification by correlated activity: Hebb's postulate revisited},
  journal = {The Journal of Neuroscience},
  volume  = {18},
  number  = {24},
  pages   = {10464--10472},
  year    = {1998},
  pmid    = {9855898}
}

@article{Gerstner1996,
  author  = {Gerstner, Wulfram and Kempter, Richard and van Hemmen, J. Leo and Wagner, Hans},
  title   = {A neuronal learning rule for sub-millisecond temporal coding},
  journal = {Nature},
  volume  = {383},
  number  = {6595},
  pages   = {76--78},
  year    = {1996},
  doi     = {10.1038/383076a0}
}

@book{Hebb1949,
  author  = {Hebb, Donald O.},
  title   = {The Organization of Behavior: A Neuropsychological Theory},
  publisher = {Wiley},
  year    = {1949},
  address = {New York}
}

@misc{kingma2017adammethodstochasticoptimization,
      title={Adam: A Method for Stochastic Optimization}, 
      author={Diederik P. Kingma and Jimmy Ba},
      year={2017},
      eprint={1412.6980},
      archivePrefix={arXiv},
      primaryClass={cs.LG},
      url={https://arxiv.org/abs/1412.6980}, 
}

\newpage

\vfill

\end{document}